# Unsupervised Anomaly Detectors to Detect Intrusions in the Current Threat Landscape

ACM TDS – Research Paper

Tommaso Zoppi

Dept. of Mathematics and Informatics, University of Florence, Italy, tommaso.zoppi@unifi.it

Andrea Ceccarelli

Dept. of Mathematics and Informatics, University of Florence, Italy, andrea.ceccarelli@unifi.it

Tommaso Capecchi

Dept. of Mathematics and Informatics, University of Florence, Italy, tommaso.capecchi@stud.unifi.it

Andrea Bondavalli

Dept. of Mathematics and Informatics, University of Florence, Italy, bondavalli@unifi.it

**ABSTRACT**

Anomaly detection aims at identifying unexpected fluctuations in the expected behavior of a given system. It is acknowledged as a reliable answer to the identification of zero-day attacks to such extent, several ML algorithms that suit for binary classification have been proposed throughout years. However, the experimental comparison of a wide pool of unsupervised algorithms for anomaly-based intrusion detection against a comprehensive set of attacks datasets was not investigated yet. To fill such gap, we exercise seventeen unsupervised anomaly detection algorithms on eleven attack datasets. Results allow elaborating on a wide range of arguments, from the behavior of the individual algorithm to the suitability of the datasets to anomaly detection. We conclude that algorithms as Isolation Forests, One-Class Support Vector Machines and Self-Organizing Maps are more effective than their counterparts for intrusion detection, while clustering algorithms represent a good alternative due to their low computational complexity. Further, we detail how attacks with unstable, distributed or non-repeatable behavior as Fuzzing, Worms and Botnets are more difficult to detect. Ultimately, we digress on capabilities of algorithms in detecting anomalies generated by a wide pool of unknown attacks, showing that achieved metric scores do not vary with respect to identifying single attacks.

• Computer systems organization → Dependable and fault-tolerant systems and networks;

• Networks;

## 1 Introduction

It is fully acknowledged that systems and networks are subject to cyber-attacks. According to the U.S.A. Committee on National Security Systems Glossary [57], **cybersecurity** is defined as "*prevention of damage to, protection of, and restoration of computers, electronic communications systems, electronic communications services, wire communication, and electronic communication, including information contained therein, to ensure its availability, integrity, authentication, confidentiality, and nonrepudiation*".

Amongst protection measures, *Intrusion Detection Systems* (IDSs, [6], [9], [13]) were proposed to enhance network and system security. IDSs collect and analyze data from networks and systems indicators to detect malicious or unauthorized activities, based on the hypothesis that an ongoing attack has distinguishable effects on such indicators.

Most enterprise IDSs adopt signature-based detection algorithms [14], [49], which search for predefined patterns (or *signatures*) in the monitored data in order to detect an ongoing attack which matches one or more signatures. Signature-based approaches usually score high detection capabilities and low false positive rates against known attacks [50], but they are not effective when the behavior of attacks gets slightly modified, calling for an update of signatures. In particular, *they are not meant to detect zero day attacks*, which are novel attacks that cannot be matched to any known signature [51]. Moreover, when a zero-day attack that exploit newly added or undiscovered system vulnerabilities is identified, its signature needs to be devised and added as a new rule to the IDS.

**Anomaly Detectors.** To mitigate the problem above, anomaly detectors are intended to *find patterns in data that do not conform to the expected behavior of a system* (or a network) [1]: these patterns are called anomalies. Anomaly-based IDS are built on the assumption that ongoing attacks will generate observable anomalies in the trend of performance indicators, or features, of the system [52] and of network [13], [50]. Unsupervised algorithms perfectly fit this activity as they can adapt themselves to suit the current context of the system and ultimately detect unknown attacks [22], [64]. However, their detection efficacy depends on the ability of characterizing the expected behavior: a poor characterization of such behavior negatively impacts on the identification of malicious anomalous activities. As a consequence, anomaly-based detectors usually generate a higher number of false alarms than signature-based methods [7], [54].

Different anomaly detection algorithms usually exhibit [24], [54] different rates of missed (False Negatives) and wrong detections (False Positives) and, consequently, have different detection capabilities. Although most of such algorithms have a generic, context-independent formulation, they are often more effective to detect specific attacks on specific systems or applications [5]. Moreover, the manifestation of the anomaly itself is usually different from attack to attack and from system to system. *Consequently, selecting the correct detection algorithm represents a crucial decision when defining an anomaly-based IDS*. An erroneous choice decreases the attack detection capabilities of the IDS, reducing its ability to secure the target system and network.

**Paper Aim.** In this paper we present a quantitative comparison of anomaly detection algorithms applied on multiple attacks datasets. We adopt unsupervised anomaly detection algorithms since they are known to be the most suitable way to deal with zero-day attacks. More in detail, we identify a total of 17 unsupervised anomaly detection algorithms that have been previously used for intrusion detection, at least one for each of the seven families (clustering, statistical, classification, neural network, neighbor-based, density-based and angle-based) that are usually considered when grouping unsupervised anomaly detection algorithms [1], [3]. Then, we identify 11 attacks datasets, namely *NSL-KDD* [18] (2009), *CTU-13* [32] (2011), *ISCX* [17] (2012), *UNSW-NB15* [16] (2015), *UGR16* [31] (2016), *NGIDS-DS* [21] (2017), *Netflow-IDS* [21] (2017), *AndMal17* [33] (2017), *CIDDS-001* [19] (2017), *CICIDS17* [20] (2017), and *CICIDS18* [20] (2018), to be used as data baseline for our experimental campaign. Since the datasets contain attacks labeled according to different nomenclatures, we also categorize the attacks of the datasets according to categories defined by the most recent studies on threat landscapes.

Exercising the selected algorithms on the datasets above allows comparing the behavior of the algorithms - individually and as families - with respect to the datasets and the attacks of our attack model. Finally, we observe how the intrinsic characteristics of attacks, along with an adequate distribution of expected and anomalous data points in the datasets helps improving detection scores, because they help algorithms to properly define the expected behavior.

This allows observing how unsupervised algorithms as Isolation Forests, One-Class Support Vector Machines and Self-Organizing Maps show less misclassifications than algorithms belonging to other families. Instead, clustering algorithm represent a valid alternative due to a good tradeoff between misclassifications and low computational complexity needed for training. Moreover, we observe that relevant attacks in the current threat landscape as Fuzzing, Worms and Botnets are trickier to detect than others, causing algorithms to generate a higher number of misclassifications with respect to attacks as web-attacks and denial of service. Detection capabilities of algorithms are also tested by submitting wide ranges of common attacks, showing that metric scores they achieve are comparable with respect to setups where the same attack was activated multiple times.

**Paper Structure.** This paper is structured as follows: Section II presents the current cyber threat landscape, zero day attacks and anomaly detection, alongside with a literature review on the comparison of anomaly detection algorithms for intrusion detection. Section III presents our experimental campaign and their inputs including the selection of the algorithms, the datasets, the attack model and the metrics that will be used for the experimental evaluation. Following sections present and discuss results: Section IV expands on algorithms, Section V focuses on detection capabilities of algorithms on each dataset and Section VI debates on detectability of specific attacks. Lastly, Section VII concludes the paper, points out potential limitations of this study and summarizes the lessons learned.

## 2 Attacks and Intrusion Detection

Security specialists are continuously looking for mechanisms and strategies that aim at neutralizing an attack or mitigating its adverse effects. Regardless of their characteristics, attacks [6], [9], [13] should be timely identified to activate reaction mechanisms that aim at blocking an ongoing attack, or protecting critical data. To such extent, many *Intrusion Detection Systems* (IDSs) were proposed in the literature to prevent attackers from exploiting security breaches or vulnerabilities. IDSs rely on monitoring activities that gather actionable data from the system under observation. The pool of system indicators that are observed e.g., resource usage, active threads, application-specific indicators, become the features to be fed to a classifier which performs intrusion detection.

In the followings we will expand on the current cyber threat landscape as described in technical reports by security agencies, and how anomaly detection may help in protecting systems, with a particular attention to the detection of zero-day attacks. To complete the section, which constitutes the baseline of the paper, we will summarize related works, positioning our paper in the literature.

### 2.1 Cyber Threats Landscape

The report from Check Point Research [10] highlights some of the most common attacks that were conducted against organizations, infrastructures and honeypots in the last two years. Authors show how malwares are the most common cyber-threat, expanding on arising trends that they expect to affect cybersecurity in the next years, namely i) *targeted ransomware*, a ransomware that is submitted to organizations through botnets, ii) *cryptominers*, which aim at exploiting the resources of the host computer to perform mining, and iii) *DNS Attacks*, which target the process of resolving domain names into their corresponding IP addresses. The rapidly growing trend of crypto-related attacks was confirmed by the last ENISA's *Threat Landscape Report* [11], which presents the most common malware families including – but not limiting to - as i) malware spread via Smartphone (mobile environment) and ii) threats to cryptocurrencies i.e., CryptoJacking and Cryptominers.

Noticeably, the ENISA [11] document provides a rank of the most common threats in the current landscape, which we report in Figure 1. Such categories are shortly described below.

- *Malware* (Rank #1 in Figure 1): a software that aims at damaging devices, stealing data, and generally threatening the system with malicious activities.
- *Web based attacks* (Rank #2): this category includes attacks that use web systems or services; the main vectors, i.e. the points and methods of entry on which these attacks can be conducted, are Browser Exploits, Drive-by-Download or malicious URLs.

- *Web Application Attacks* (Rank #3): they exploit the software vulnerabilities of web applications e.g., SQL-injection (SQLi). Their increasing relevance is due to digital applications and services that have been – and are currently being - ported on web services.
- Web/Mail Attacks (*Phishing* #4, and *Spam* #6): Spam is the abusive use of email and messaging technologies to flood users with unsolicited messages, while phishers try to lure the recipients of phishing emails and messages to e.g., open a malicious attachment, click on an unsafe URL, hand over their credentials via legitimate looking pages, wire money. Over 90% of malware infections in organizations originate from phishing attacks [12].
- *Denial of Service* (DoS, Rank #5): DoS and DDos (Distributed DoS) attacks represent a category of attacks that damage availability of platforms or services. Common threats exploit vulnerabilities of the UDP, TCP and ICMP network protocols.
- *Botnets* (Rank #7): a network of Bots (computers compromised by malware - zombies), coordinated by a malicious Botmaster. Despite this type of attack is often associated with DoS this type of attack is used in various areas: financial-economic, Internet of Things and Cryptocurrency.
- Data Breaches (Rank #8): this category groups all threats to integrity and/or confidentiality of data. It includes all the scanning, probing and reconnaissance attacks that passively – and, often, stealthily – gather information about a given system, organization or individual, and specific types of malware who steal data from the victim.

Note that we stopped at the 8<sup>th</sup> position in the rank, because from that point on the report [11] starts describing threats which may not have clear impacts on system data, as *insiders* or *physical damages*. These are primarily meant to be mitigated or avoided through mechanisms other than intrusion detection i.e., access control, physical protections to components.

| Top Threats 2017 | Assessed Trends 2017 | Top Threats 2018 | Assessed Trends 2018 | Change in ranking |
|---|---|---|---|---|
| 1. Malware | Stable | 1. Malware | Stable | → |
| 2. Web Based Attacks | Increasing | 2. Web Based Attacks | Increasing | → |
| 3. Web Application Attacks | Increasing | 3. Web Application Attacks | Stable | → |
| 4. Phishing | Increasing | 4. Phishing | Increasing | → |
| 5. Spam | Increasing | 5. Denial of Service | Increasing | ↑ |
| 6. Denial of Service | Increasing | 6. Spam | Stable | ↓ |
| 7. Ransomware | Increasing | 7. Botnets | Increasing | ↑ |
| 8. Botnets | Increasing | 8. Data Breaches | Increasing | ↑ |
| 9. Insider threat | Stable | 9. Insider Threat | Declining | → |
| 10. Physical manipulation/ damage/ theft/loss | Stable | 10. Physical manipulation/ damage/ theft/loss | Stable | → |
| 11. Data Breaches | Increasing | 11. Information Leakage | Increasing | ↑ |
| 12. Identity Theft | Increasing | 12. Identity Theft | Increasing | → |
| 13. Information Leakage | Increasing | 13. Cryptojacking | Increasing | NEW |
| 14. Exploit Kits | Declining | 14. Ransomware | Declining | ↓ |
| 15. Cyber Espionage | Increasing | 15. Cyber Espionage | Declining | → |

Legend: Trends: ↓ Declining, ⊃ Stable, ↑ Increasing
Ranking: ↑ Going up, → Same, ↓ Going down

Figure 1. Common threats from ENISA's Threat Landscape [11].

## 2.2 Zero-Day Threats

The wide range of cyberattacks, alongside with their natural ability to evolve, obfuscate and hide in between legitimate events, make them difficult to understand and analyze. However, antiviruses and enterprise IDS embed signature-based approaches to detect known faults [14], [13]. Briefly, a *signature* – or *fingerprint* – of each known attack is derived and added to a local database, which checks if the state of the system matches with at least one signature of the known threats. However, a solid IDS cannot rely only on detecting known – and expected – attacks; therefore, security specialists should consider mechanisms that are able to detect zero-day attacks [51], [54]. These mechanisms are not meant to replace traditional signature-based approaches that were proven to be effective in detecting known threats. Instead, they should be used alongside with traditional mechanisms to provide a complete protection against intrusions.

To deal with unknowns, research moved to techniques suited to detect unseen, novel attacks. Anomaly detectors [1] are based on the assumption that an attack generates observable deviations from an expected – normal – behavior. Briefly, they aim at finding patterns in data that do not conform to the expected behavior of a system: such patterns are known as anomalies. Once an expected behavior is defined, anomaly detectors target deviations from such expectations, protecting against known attacks [2], [7], zero-day attacks [6] and emerging threats [8]. To such extent, most of the anomaly detection algorithms are *unsupervised*, suiting the detection, among others, of zero-day attacks, without requiring labels in training data [4], [3], [5].

## 2.3 Unsupervised Anomaly Detection Algorithms and Families

In the paper we refer to *data point* as the observation of the state of the system at a given instant, which is described by an item in *datasets*. Each data point is composed by *f feature values*, which are processed by an anomaly detection algorithm to determine if the data point exhibits anomalies.

More in detail, anomalies are rare data points that were classified [1] as: i) *point* anomaly (global outlier): a data point that is out of scope or not compliant with the trend of a variable e.g., out-of-size payload of a network packet; ii) *contextual* anomaly (local outlier): a data point that is unexpected in a specific context e.g., low number of page faults while loading a program for the first time, or iii) *collective* anomaly: a collection of related data points that is anomalous with respect to the entire trend or dataset e.g., subsequent ICMP requests in a short interval of time.

Different anomaly detectors may be instantiated depending on the nature of the target system and monitored data. If labeled training data is available, supervised anomaly detection or semi-supervised may be adopted [2]. Labelled data points allow training an algorithm using both expected and anomalous data points that have already been reported. This way, the algorithm is fed with anomalies due to known attacks and learns how they differ from expectations, disregarding the detection of anomalies due to unseen attacks. Instead, when focusing on zero-day attacks, or when training data is not labeled, the only option is an unsupervised anomaly detection approach [4], [3], [5], [54].

Different unsupervised anomaly detectors have been proposed throughout years. They rely on ML algorithms that are suited for binary classification, to distinguish between normal and anomalous data points. Throughout years, unsupervised algorithms have been studied and compared [64] to derive similarities or differences, devising families of algorithms [1], [3], [5], [54]. *Clustering* algorithms partition a dataset grouping data points in the same cluster if they share similar characteristics. Data points that cannot be assigned to any of the existing clusters, or that do not meet specific inclusion criteria, are anomalous. Instead, n*eighbor-based* algorithms label a data point as anomalous or expected depending on the distance with respect to its nearest neighbor(s). Similar algorithms, instead, estimate the *density* of the surroundings of each data point. An alternative detection mechanism is provided by *angle-based* algorithms, which measure the variance in the angles between the data point to the other known data points; anomalies typically result in very small variance of angles. *Classification* algorithms identify the binary class of a new data point devising adequate boundaries, while s*tatistical* algorithms assume that anomalous data points occur in low probability regions of a given statistical distribution derived during training. Lastly, unsupervised *neural networks* produce a two-dimensional, discretized representation of the input space, the so-called map. It is worth noticing that there are some unavoidable semantic overlaps among families; for example, nearest-neighbour methods may be used to define variations of algorithms as in the angle-based *FastABOD* [26].

## 2.4 On the Comparison of Unsupervised Algorithms

In such a context, we think that a deep comparison among anomaly detection algorithms for intrusion detection is needed to understand which (family of) algorithm is recommended when dealing with attacks and systems. Throughout years, multiple comparison studies were proposed [5], [4], [3]. The authors of [58] used 7 algorithms on a single proprietary dataset containing HTTP traffic, providing an open-source IDS testing framework. Similarly, in [59] authors evaluate 4 algorithms on a single dataset, focusing more on feature selection. Instead, in [7], authors presented a comparative study for intrusion detectors where k-Nearest Neighbors (kNN), Mahalanobis-based, Local Outlier Factor (LOF) and one-class Support Vector Machines (SVM) were evaluated using only the DARPA 98 dataset and real network data (for a total of 2 datasets). Similarly, in [61] authors compared three unsupervised anomaly detection algorithms for intrusion detection: Cluster-based Estimation, kNN and one-class SVM using network records stored in the KDD Cup 99 dataset and system call traces from the 1999 Lincoln Labs DARPA evaluation. Four algorithms are evaluated in [60], which presents a review of novelty detection methods that are classified into semi-supervised and unsupervised categories. The algorithms are exercised on 10 different datasets regarding medical and general-purpose data.

Summarizing, it is not easy to find studies that apply a wide range of unsupervised algorithms to datasets containing data related to the most common attacks in the current threat landscape. In [58] the authors considered a single proprietary dataset, while the work in [7] uses two datasets and four algorithms, without taking into account all the main families of algorithms defined in [1] and refined in [3]. Similarly, in [61] the authors used 3 algorithms on 2 datasets, while, [2] uses 3 unsupervised algorithms on 2 datasets. Only [5] executes different algorithms on a (small) group of datasets and organizes the results according to a unified attack model. All these studies do not provide a comprehensive view on algorithms to be applied for unsupervised anomaly detection, and execute experiments without adopting a shared methodology. Therefore, *we strongly believe that a quantitative comparison of unsupervised anomaly detection algorithms applied on multiple attacks datasets can provide researchers and practitioners a solid baseline on detection capabilities of multiple algorithms when detecting current threats*.

## 3 Quantitative Analyses and Experimental Campaign

To substantiate and elaborate on how and if different unsupervised anomaly detection algorithms can detect attacks in the current threat landscape, it is fundamental to quantitatively evaluate their detection capabilities. Therefore, we planned and executed an experimental campaign as follows.

- We collect public datasets which contain data related to the attacks in Section 2.1 [11]. Datasets should be recent and collected while monitoring real systems.
- Then, we review the literature aiming at finding unsupervised algorithms which are meant to perform binary classification and are therefore suitable for anomaly detection. Amongst all the possible alternatives, we disregard variants of algorithms, aiming instead to span across all families identified in Section 2.3.
- Then, we apply each algorithm to each dataset, collecting metric scores. These metrics describe detection capabilities of algorithms on each datasets, accounting also for classification errors, which are relevant items to discuss results.
- Once the experiments have been executed, discussion on experimental data should explore i) detection capabilities of algorithms across all datasets/attacks, ii) attacks or categories of attacks that are particularly easy/difficult to identify, and iii) datasets that turn out to be particularly easy or challenging for anomaly detection.

Different inputs are needed to execute our experimental campaign. First, Section 3.1 describes publicly available datasets that contain data about normal behavior of a system and attacks relevant to the current threat landscape. Then, unsupervised algorithms are briefly introduced in Section 3.2, leaving Section 3.3 to describe the metrics that will be used to evaluate detection capabilities of algorithms on datasets. The supporting tool and further details on the implementation of the experimental campaign are summarized in Section 3.4 and Section 3.5, which complete the digression on the experimental campaign and makes room for discussions in the rest of the paper.

### 3.1 Publicly Available Datasets

Attack datasets are usually structured as a set of features that describe the system indicators being monitored by system owners. They contain a given amount of data points, which are observation of system indicators i) when specific events happen, or ii) at predefined time intervals. Moreover, in most of the cases authors provide labels to group data points that were collected when system was under attack. While *these labels are not needed to train unsupervised algorithms, they turn out useful to estimate detection* *capabilities of algorithms through confusion matrix-based metrics*, which also allow comparing the effectiveness of different algorithms on the same dataset.

Starting from [15] and by querying online portals[1,2], we select datasets with the following characteristics: i) published recently, ii) labelled (at least partially), iii) that contain at least one of the attacks in the ENISA top 10, and iv) already used in the literature for intrusion detection studies. The resulting datasets are shown in Table 1 and briefly described below.

- *(2009) NSL-KDD* [18]. This dataset solves problems in the KDD Cup 99 dataset as i) the presence of redundant records in train sets, and ii) duplicates in test sets. The dataset contains the following attacks: DoS, R2L (unauthorized access from a remote machine), U2R (unauthorized access to super-user or root functions) and Probing (gather information about a network).
- *(2011) CTU-13* [32]. The CTU-13 is a dataset of botnet traffic that was captured in the CTU University, Czech Republic, in 2011. The goal of the dataset was to have a large capture of real botnet traffic mixed with normal traffic and background traffic. The CTU-13 dataset consists in thirteen captures (called scenarios) of different botnet samples.
- *(2012) ISCX12* [17]. It is generated in a controlled environment based on a realistic network and traffic, to depict the real effects of attacks over the network and the corresponding responses of workstations. Four different attack scenarios are simulated: infiltration, HTTP denial of service, a DDoS by using an IRC botnet, and SSH brute-force login attempts.
- *(2015)UNSW-NB15* [16]. Released by the Australian Defense Force Academy in the University of New South Wale, it contains: i) Exploits of a generic vulnerability, ii) DoS, a (Distributed) Denial of Service, iii) Worms, iv) Fuzzers, v-vi) Reconnaissance and Analysis, attacks that aim at gathering information, vii) Shellcode, a code used as the payload in exploits, and viii) Backdoors, to bypass security mechanisms and access sensitive data.
- *(2016) UGR16* [31]: UGR is built with real traffic and up-to-date attacks. These data come from several *netflow v9 collectors* strategically located in the network of a Spanish ISP. The dataset considers long-term evolution and traffic periodicity, and embeds normal traffic as well as data related to DoS, BotNet, Scan, Blacklist and Spam attacks.
- *(2017) NGIDS-DS* [21]. It contains network traffic in packet-based format as well as host-based log files. It was generated in an emulated environment, using the IXIA Perfect Storm tool to generate normal user behavior as well as attacks from seven different attack families (e.g. DoS or worms).
- *(2017) Netflow-IDS* [21]: the dataset was created at the next generation cyber range infrastructure of the Australian Centre OF Cyber Security (ACCS) for the Australian Defence Force Academy (ADFA), Canberra. It is part of the ongoing projects in the ADFA related to the cyber security. It contains normal and abnormal host (LINUX) and network activities which are performed during the emulation. It contains DoS attacks as Neptune, Portsweep, and MailBomb (Spam), which authors deemed relevant for future IDS design.
- *(2017) AndMal17* [33]. The dataset was collected by running both malicious and benign applications on real smartphones to avoid runtime behavior modification of advanced malware samples that are able to detect the emulator environment. Data relates to over six thousand benign apps from Google Play market published in 2015, 2016, 2017, and contains the following macro-categories: Ransomware, Scareware, SMS Malware, Adware.
- *(2017) CIDDS-001* [19]. The CIDDS-001 data set was captured within an emulated small business environment in 2017, contains four weeks of unidirectional flow-based network traffic, and contains several attacks captured from the wild.
- *(2017) CICIDS17* [20]. CICIDS 2017 was created within an emulated environment over a period of 5 days and contains network traffic in packet-based and bidirectional flow-based format. For

---

[1] AZSecure – Intelligence and Security Informatics Datasets, https://www.azsecure-data.org/other-data.html

[2] UNB – Canadian Institute for CyberSecurity, https://www.unb.ca/cic/datasets/index.html

Table 1. Mapping of Attacks to Datasets. Overall, 51 combinations of <dataset, attack> are shown.

| Attack Category<br>ENISA Rank | Malware<br>1 | Web Attack<br>2 | Web Application<br>3 | Spam / Phishing<br>4, 6 | (D)Dos<br>5 | BotNet<br>7 | Data Breaches<br>8 |
|---|---|---|---|---|---|---|---|
| *NSL-KDD* | u2r | | r2l | | DoS | | Probe |
| *CTU-13* | | | | | | BotNet | |
| *ISCX12* | | BruteForce | | | DoS, DDoS | | Infiltration |
| *UNSW-NB15* | Worms | Fuzzers | Backdoor, Exploits, Shellcode | | DoS | | Analysis, Reconnaissance |
| *UGR16* | | | | Blacklist, Spam | DoS | BotNet | Scan |
| *NGIDS-DS* | Malware, Worms | | Backdoor, Exploits, Shellcode | | DoS | | Reconnaissance |
| *Netflow-IDS* | | | | Mailbomb | Neptune, Portsweep | | |
| *AndMal17* | Ransomware, Scareware | | | SMS, Adware | | | |
| *CIDDS-001* | | BruteForce | | | DoS | | PortScan, PingScan |
| *CICIDS17* | | BruteForce | | | DoS (Slowloris, Goldeneye) | | PortScan |
| *CICIDS18* | | BruteForce (FTP, SSH) | | | DoS, DDoS | Bot | Infiltration |

  each flow, the authors extracted more than 80 features. The data set contains a wide range of attack types like SSH brute force, Botnet, DoS, DDoS, web and infiltration attacks.
- *(2018) CICIDS18* [20]. Similarly to CICIDS 2017, CICIDS 2018 was created as an updated version of the previous dataset, containing Brute-Force, Botnet, DoS, DDoS, Web i.e., SQLi, and Infiltration attacks.

Overall, the 11 datasets we selected are mapped with respect to ENISA attacks as shown in Table 1. The table allows identifying 51 couples of <*dataset, attack*>, pointing to portions of datasets logging a single type of attack and normal behavior. Datasets as CICIDS17 or AndMal17 are created by exercising specific attacks in different timeframes and already report on separate files containing normal behavior and attack data. Other datasets as NSLKDD or UGR16 log different attacks in the same timeframe, still without overlaps i.e., at most one attack is conducted at a given time. In the latter case, we skipped data points reported on attacks other than he target while building portions of the datasets. It is worth noticing that each of the 8 most common attack categories is mapped onto at least 3 different datasets: as a result**, the datasets above are representative of the most common attacks of the current threat landscape**.

## 3.2 Unsupervised Algorithms

As shown in Figure 2, we choose a heterogeneous set of 17 different unsupervised algorithms that are grouped according to the family they belong and are briefly described below. The heterogeneous choice of algorithms allows embracing a **wide range of algorithms that build on different heuristics, giving this study a solid algorithm baseline** to rely upon. Note that Neural Network, Classification and Neighbour-based algorithms have a computational complexity that is at least quadratic, while the angle-based computation employed by ABOD [26] has cubic complexity. Clustering and Statistical algorithms, instead, may have semi-linear computational complexity i.e., [25], [34], [35].

### 3.2.1 Clustering (K-Means, G-Means)

*K-means* [44] is a popular clustering algorithm that groups data points into k clusters by their feature values. First, the k cluster centroids are randomly initialized. Then, each data record is assigned to the cluster with the nearest centroid, and the centroids of the modified clusters are re-calculated. This process stops when the centroids are not changing anymore. Scores of each data point inside a cluster are calculated as the distance to its centroid.

Stemming from K-Means, *G-Means* [34] automatizes the choice of the optimal k through subsequent repetitions of trainings, assuming that data points inside clusters are distributed according to a normal distribution.

### 3.2.2 Statistical (HBOS, SOS)

The *Histogram-based Outlier Score* (HBOS) algorithm is a statistical approach [25] that generates an histogram for each independent feature of the given dataset. The values of the features of all the available data points are first used to build histograms; at a later stage, for each data point, the anomaly score is calculated as the multiplication of the inverse heights of the columns in which each of its features falls.

Instead, Stochastic Outlier Selection (SOS, [43]) employs the concept of affinity to quantify the relationship from one data point to another data point. Affinity is proportional to the similarity between two data points: a data point has little affinity with a dissimilar data point. As a result, a data point is more likely to be anomalous when all the other data points have insufficient affinity with it.

### 3.2.3 Classification (One-Class SVM, Isolation Forests)

The *One-class Support Vector Machine* (one-class SVM) algorithm aims at learning a decision boundary to group the data points [28]. It can be used for unsupervised anomaly detection, despite at first supervised support vector machines (SVMs) were used only for (semi-)supervised anomaly detection. The one-class SVM is trained with the dataset and then each data point is classified considering the normalized distance of the data point from the determined decision boundary.

*Isolation Forest* (IF) [42] is built as an ensemble of *Isolation Trees*. Each isolation tree structures data points as nodes, assuming that anomalies are rare events with feature values that differ a lot from expected data points. Therefore, anomalies are more susceptible to isolation than the expected data points, since they are isolated closer to the root of the tree instead of the leaves. It follows that a data point can be isolated and then classified according to its distance from the root of the trees of the forest.

### 3.2.4 Neural Network (Self-Organizing Maps)

Kohonen's *Self-Organizing Maps* (SOM) were proposed [30] as an unsupervised alternative to most of the neural-networks based classifiers. More in detail, it is an artificial neural network that is trained using unsupervised learning to produce a binary representation of the input space of the training samples i.e., a map. Self-organizing maps differ from other artificial neural networks as they apply competitive learning - through a neighborhood-like function - as opposed to error-correction learning i.e., backpropagation with gradient descent.

Figure 2. Unsupervised Algorithms Selected for this study, grouped through families.

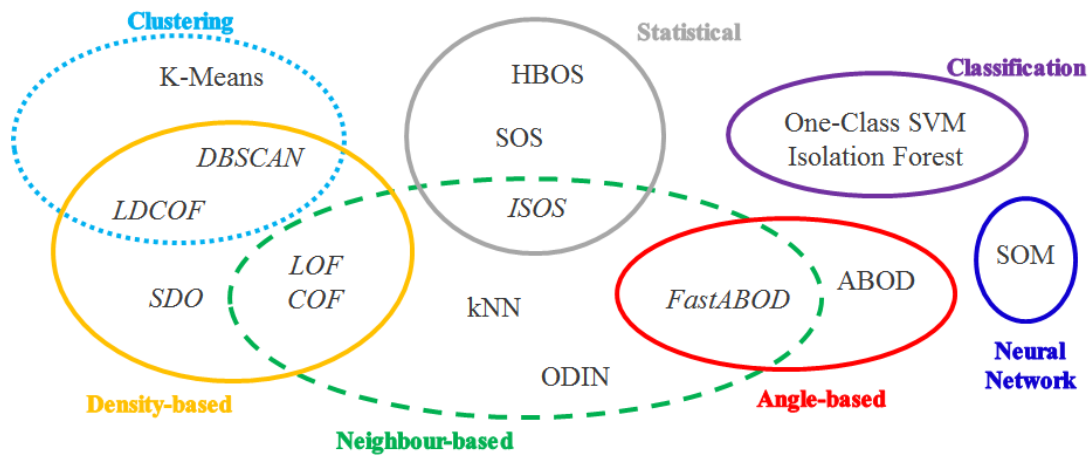

### 3.2.5 Angle-Based (ABOD)

*Angle-Based Outlier Detection* (ABOD) [26] relates data to high-dimensional spaces, using the variance in the angles between a data point to the other points as anomaly score. Each data point in the dataset is used as the middle point p2 of a polygonal chain (p1,p2,p3), while p1 and p3 are any two different data points of the dataset, p1 , p2 , p3. Then, all the angles p1p2p3 are measured, and their variance is used to calculate the Angle-Based Outlier Factor (ABOF). Ultimately, anomalies typically result in very small variance in the angles from couples of points.

### 3.2.6 Neighbour-Based (kNN, ODIN)

*Kth-Nearest Neighbor* (kNN) [40] is a neighbour-based method which was primarily designed to identify outliers in a supervised fashion. For each data point, the whole set of data points is examined to extract the k items that have the most similar feature values: these are the k nearest neighbors (NN). Then, the data point is classified as anomalous if the majority of NN was previously classified as anomalous. Note that the nearest neighbours are gathered in an unsupervised manner: therefore, unsupervised variants [41] of this method use the distance to the k-th neighbour as anomaly score.

*Outlier Detection using Indegree Number* (ODIN) [27] stems from kNN, which examines the whole dataset to determine their feature distances to the given point. This allows isolating NN, creating the kNN graph. Differently from kNN, ODIN classifies as anomalies the data points that have a low number of in-adjacent edges in the kNN graph.

### 3.2.7 Density-Based (Sparse Density Observers)

SDO (*Sparse Data Observers*) was devised [29] to detect anomalies based on low density models of data with reduced computational costs. Briefly, it labels specific items of the training set as observers, which are then used to calculate the anomaly score as the average distance of a data point to each observer. The initial choice of the observers is randomized, and then refined to identify the most meaningful observers.

### 3.2.8 Algorithms with Semantic Overlaps (LOF, COF, DBSCAN, LDCOF, ISOS, FastABOD)

To complete the selection process, we choose other 6 algorithms which have cross-cutting peculiarities among families. Neighbors identification is employed to reduce noise and computational complexity in the stochastic ISOS [37], the angle-based FastABOD [26] and in the density-based LOF [39] and COF [36]. Interesting mixtures of clustering and density-based families allow devising DBSCAN [38] and LDCOF [35], which builds a density-based anomaly detector on top of an internal clustering procedure.

## 3.3 Evaluation Metrics

The effectiveness of anomaly detectors is usually assessed using correct detections (true positives TP, true negatives TN) and wrong detections (false negatives FN, false positives FP), which build the so-called *confusion matrix*. As in [22], aggregated metrics as Precision, Recall (or Coverage), False Positive Rate, Accuracy, F-Score($\beta$), F-Measure (F1), Area Under ROC Curve (AUC, [24]) and Matthews Coefficient (MCC, [23]) are used in different studies, depending on the domain. FP-inclined metrics as Precision, FPR and F-Score (with $\beta < 1$), are relevant when the number of false alarms needs to be as low as possible e.g., to increase usability, while Recall and F-Score (with $\beta > 1$) are more relevant in those systems where FNs may constitute severe threats e.g., safety-critical systems, which also need domain-specific metrics [62].

For the sake of generality and easiness of comparison with other studies, in this study we mainly report on widely used metrics that weight FPs and FNs as equally undesired i.e., F-Measure, Accuracy, MCC. However, with unbalanced datasets i.e., if a file reports on many attacks data and a few normal data, some of the metrics above can be misleading [56], since they either i) do not consider all the four classes of the confusion matrix i.e., F1, FScore(β), or ii) consider all the classes without weighting the size of trues and falses i.e., Accuracy. To such extent, in this paper we mainly discuss *MCC*, which does not show the weaknesses above.

## 3.4 Tool Support

To execute experiments, we account for tools that allow: i) loading datasets related to attacks, mainly CSV and ARFF files, ii) executing unsupervised algorithms, and iii) extracting the above mentioned metrics. Amongst all the available frameworks as ELKI[3], WEKA[4] or libraries as Pandas[5], our final choice has been RELOAD [45], an open-source tool[6] that wraps the implementation of several unsupervised algorithms, that are often deemed the most useful [3], [7], [5] for unsupervised anomaly detection in cyber-security. Moreover, the tool allows running experiments through an intuitive user interface, automates the selection of the most relevant features out of a dataset, embeds automatic tuning of algorithms' parameters, includes built-in metrics for the evaluation and facilitates examining outputs through CSV files and graphical plots. Metrics are calculated by RELOAD in two steps: i) a data point is processed by an algorithm, which outputs a numeric score, then ii) a threshold – or decision function [63] - is applied to convert numeric scores into boolean. This threshold is defined during training by the tool through grid search amongst variants of interquartile range or confidence interval functions. Note that this tool alone allows running all the algorithms selected for this study: relying on a single tool allows minimizing possible errors in the preparation and formatting of inputs/outputs.

## 3.5 Experiments Setup and Execution

We describe here the experimental setup for our study, which is also summarized in Figure 3.

**Datasets/Tool Download.** We downloaded the datasets in Section IV.A from their repositories shaping them as CSV files. Then, we downloaded the latest release of RELOAD, setting up its parameters.

**Metric Setup.** We adopt MCC as target metric: while this metric is used by RELOAD to find optimal parameters values of algorithms, metrics other than MCC are still reported as output and will be discussed in the remaining of the paper.

**Feature Selection.** According to literature studies, out of the feature selection strategies made available by RELOAD, we choose Information Gain [46] as *feature selection* strategy. Then, the tool allows either to select the features who reach or exceed a given information gain score, or selecting the $n$ features that get the highest information gain rank. Since the former strategy may result in a different number of selected features for each dataset, we choose to adopt the latter option RELOAD provides, to standardize the amount of selected features across all the datasets. Since several datasets have just a few numeric – non categorical – features, we set $n = 3$.

**K-Fold.** We proceed with a 10-fold sampling of the training set as widely suggested in literature [47].

**Choice of Algorithms Parameters.** Besides algorithms as ABOD and G-Means, which do not rely on parameters, we try combinations of parameters e.g., $k$ for kNN-based algorithms k ϵ {1, 2, 3, 5, 10, 20, 50, 100}, and select the value which allows obtaining the best MCC value in a small portion of dataset (not overlapping with the evaluation set) which is used for testing. We consider this range as acceptable, since the amount of data points employed for training is always in the 5.000 – 10.000 range, lower for algorithms as ABOD that have cubic complexity and therefore escalate into memory errors when using many data points.

---

[3] "Elki data mining," elki-project.github.io, accessed: 2019-11-20

[4] "Weka 3: Data Mining Software in Java", www.cs.waikato.ac.nz/~ml/weka/, accessed: 2019-11-20

[5] McKinney, Wes. Python for data analysis: Data wrangling with Pandas, NumPy, and IPython. " O'Reilly Media, Inc.", 2012.

[6] RELOAD Wiki, GitHub, https://github.com/tommyippoz/RELOAD/wiki, 2020-01-17

Figure 3. Building blocks of the Experimental Setup and Methodology used in this paper.

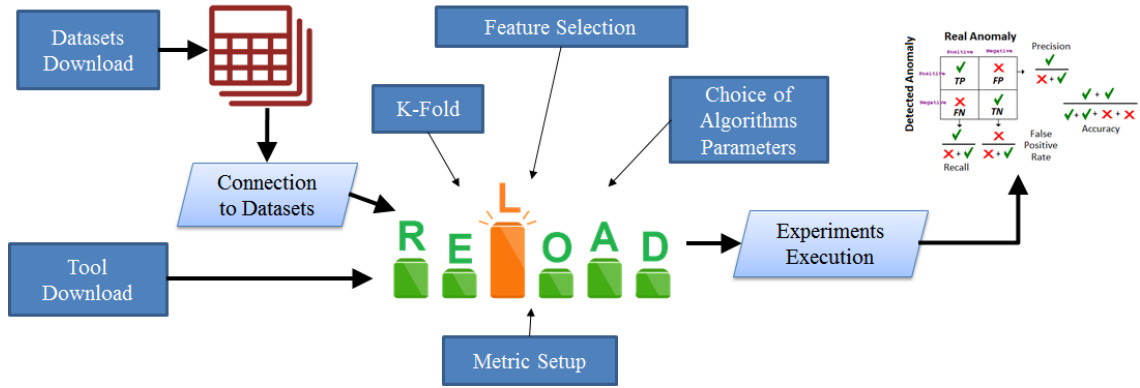

**Connection to Datasets.** RELOAD allows defining loaders i.e., data files which specify parameters to gather data from different data sources. We create a loader for each combination of <dataset, attack>; each loader points to two *separate* training and validation sets. The validation set should contain at least one data point which corresponds to the given attack, to calculate metrics scores.

**Machine to Execute Experiments.** Once all the parameters above are set, we run experimental campaigns including all the datasets and algorithms considered in this study. The experiments have been executed on a server equipped with Intel Core i7-6700 with four 3.40GHz cores, 24GB of RAM and 1TB of user storage.

**Experiments Execution.** Overall, executing the experiments required approximately 25 days of 24H execution. We executed all the 17 algorithms on all the datasets specified by the resulting 51 loaders, collecting confusion matrixes and metric scores computed by RELOAD. Overall, we obtained a total of 867 triples <*algorithm, loader, metric_values_on_validation*>, which are going to be presented and discussed in the next sections.

Moreover, to sustain the analysis we will be presenting in Section 6.3, we executed additional experiments by training algorithms once for each dataset, using a training set which contains a mixture of normal data points and *all* the attacks considered in the datasets. While some datasets as NK, NG, UG provide single files which already contain this mixture, we were forced to artificially create it for other datasets, which report logs of different attacks executed in different days and stored as separate files. This resulted in additional 187 experiments generating triples structured as above, where each of our 17 algorithms was run on each of the 11 datasets.

All the metric scores and files that we used to collect and summarize values are publicly available at [55].

## 4 Results of Unsupervised Anomaly Detectors and Families

After defining the experimental campaign, its inputs, and the way it has been exercised, in this section we start reporting experimental results, alongside with discussions. *All the experimental results, parameters setup, and intermediate RELOAD data, including metric scores that cannot be shared in this document for brevity, are publicly available at [55].*

### 4.1 Detection Capabilities of Algorithms

We first explore the detection capabilities of each algorithm selected in this study. Our experimental campaign provides metric scores for each algorithm on each of the 51 couples <dataset, attack>: these 51 results can be aggregated to obtain the results in Figure 4. In the figure we report 4 different series of bars representing i) average of MCC scores, with error bars reporting on standard deviation, ii) average F1, iii) average ACC, and iv) # Best MCC, which counts the amount of datasets in which a given algorithm shows the best MCC.

Figure 4. Results of *MCC*, *F1*, *ACC* for each algorithm considered in this study, aggregated through datasets and sorted left to right according to a decreasing MCC. On the secondary axis the plot shows *# Best MCC*, which shows the amount of datasets in which a given algorithm shows the best MCC for that specific dataset.

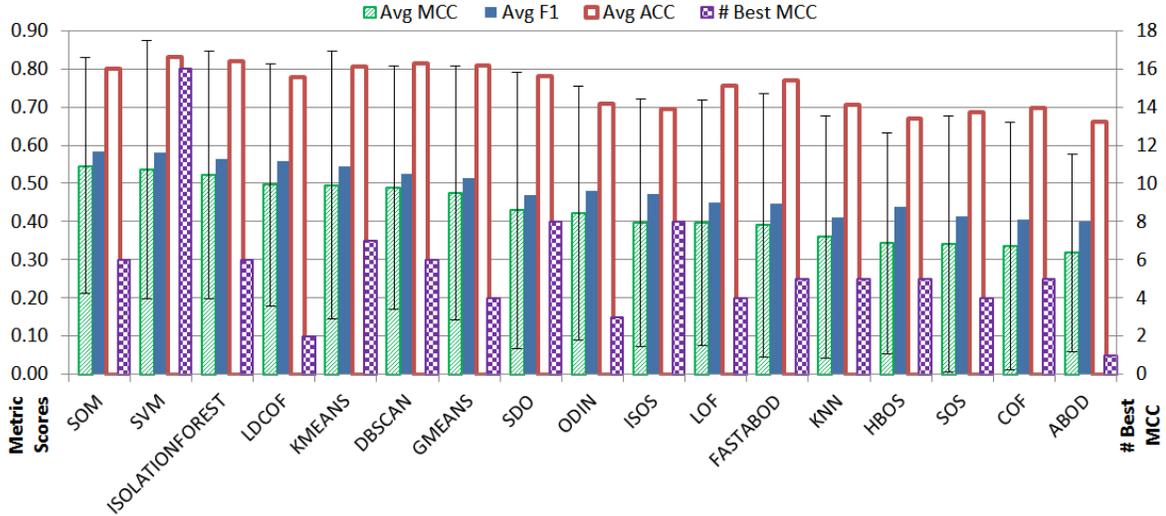

This view on results allows elaborating on individual capabilities of algorithms, which are shown from left to right according to their decreasing average MCC scores, depicted as green-striped bars. Algorithms on the left of the figure can be considered as the best choices according to the average value of MCC they provide. Noticeably, the average MCC alone does not provide enough detail to substantiate our analysis, given the variability of scores as highlighted by error bars reporting on standard deviation. Therefore Figure 4 reports also on metrics other than MCC i.e., F1 and Accuracy, which are broadly used in similar studies. F1 follows a trend which is similar to MCC: in the general case, high values of MCC are usually paired with high values of F1. Instead, Accuracy shows high values also for algorithms as LOF, FastABOD, which do not have high MCC and F1 average values. Accuracy values are relatively high due to an amount of FNs and FPs which is low with respect to the amount of TPs and TNs. However, with unbalanced datasets high values of accuracy may not be informative as other metrics [56] which aggregate the confusion matrix differently.

*The purple-squared bar series in Figure 4 reports on the amount of datasets where a given algorithm shows the highest MCC scores out of the pool of available algorithms*. Algorithms with very similar values of MCC, F1 or ACC may have a different ranking according to this metric. An algorithm may be either "good on average", or being very good in specific situations and not so competitive in other context. This is the case of SOM and One-Class SVM. Both show the best average MCC scores i.e., 0.54 among all the 51 couples of datasets and attacks, but have very different values of this last metric. In particular, **SVM is the best algorithm for 16 out of the total 51 <*dataset, attack*> (31.3%),** while SOM shows excellent detection capabilities only in 5 situations. Standard deviation of MCC scores results in 0.29 of SOM with respect to 0.35 for SVM, as it is confirmed by shorter error bars for SOM in Figure 4. Consequently, *SVM has more variability in its results*: this leads to more situations in which it is the best algorithm, but exposes to other situations in which SVM may raise more FNs and FPs with respect to most of the other algorithms.

Similarly to SVM, SDO shows high variability of MCC scores, reaching a standard deviation of 0.36, the highest out of all algorithms, even higher than SVM. While its average MCC does not stand out, *SDO is the optimal algorithm in 8 couples of dataset-attack, more than SOM*, which has a higher average MCC (0.54, against the 0.43 of SDO). **Huge variability of MCC scores provided by algorithms point out that they may either be very good or very bad depending on the dataset**.

## 4.2 Focusing on Algorithm Families

Starting from the results of individual algorithms, we can observe our experimental results according to algorithms families. As described in Section 2.3, each algorithm belongs to at least one family, while semantic overlaps of an algorithm between two or more families cannot be avoided. In these cases, algorithms are considered as belonging to all the overlapping families. Figure 5 reports data aggregated with respect to families. When families appear as surrounded by * on the left of the bar chart, the values expressed by bars are calculated considering all the algorithms that belong to that family, including those that overlaps with other families. Instead, when the name of the family appears without special characters, data is obtained considering scores of algorithms that belong only to the specific family. The bar chart in Figure 5 reports the average MCC as well as the average *Rank MCC*. This metric is obtained as follows: for each dataset, we rank each of the 17 algorithms by assigning a rank 16 to the algorithm with the best MCC for that dataset, going down to rank 0, which is assigned to the algorithm which resulted in the worse MCC score for that dataset. Consequently, the higher the average Rank MCC, the more frequently a given algorithm is a top choice.

**Angle-Based and Statistical Families.** Families in Figure 4 are ordered from top to bottom according to an increasing average MCC. Lowest scores are obtained by ABOD, which is the only angle-based algorithm that does not overlap with other families. Since ABOD was proven to be effective in different studies as [26], we motivate the mediocre scores considering that we were forced to train ABOD on smaller sets with respect to what it happens with other algorithms due to its cubic time complexity. This lack of data points in the training set may have impacted detection capabilities of the algorithm, resulting in low MCC scores.

Statistical algorithms come right after ABOD in Figure 5: despite they are meant to be very fast as they execute training in linear time, sometimes they are not able to adequately fit train data into any probabilistic distribution, resulting in an higher amount of misclassifications with respect to other algorithms.

**Neighbour and Density-Based.** It is worth noticing the role of neighbor-based algorithms, which constitute the majority of algorithms we selected for this study. In addition to kNN and ODIN, there are 4 additional algorithms i.e., LOF, COF, ISOS, FastABOD, that embed a kNN search to reduce their computational complexity and therefore need to be considered as neighbor-based strategies. These algorithms do not reach MCC values that are as good as other families as Classification (SVM, Isolation Forests) or Neural Networks (SOM). This conclusion has a large impact on our study, since these widespread algorithms show average MCC of 0.38 (0.39 when considering overlaps) that are far lower with respect to their counterparts as classification (MCC of 0.53) and SOM (0.54). The Rank MCC varies accordingly: 8.1 (8.2) for neighbor family, and 11.2 and 11.1 respectively for Classification and Neural Networks (SOM). To a lesser extent, these conclusions translate to density-based algorithms, which usually build a density index upon a neighbor or cluster search. Overall scores of the family are slightly better than neighbor-based algorithms. However, there is a clear separation between scores of density algorithms based on kNN i.e., LOF, COF, and others i.e., SDO, DBSCAN, LDCOF, which have better scores (see Figure 4).

It follows that **density and neighbor-based algorithms, despite their widespread distribution, should not be chosen blindly as intrusion detectors, as they do not usually represent the best alternative out of available unsupervised algorithms**. This problem was already acknowledged in the literature [1], [53], leading research to lean towards other approaches as angle-based or clustering, which indeed have their own weaknesses.

**Clustering.** Sticking on clustering, Figure 5 shows how these algorithms *have overall convincing detection capabilities, immediately following SVM, Isolation Forests, and SOM algorithms, with the possibility of executing training in non-quadratic time*. Aside of DBSCAN, K-Means, G-Means and LDCOF all work in sub-quadratic time and show good average MCC scores. As a cross-check, clustering algorithms hold positions 4 to 8 in Figure 4, which highlights individual scores of algorithms. This opens an interesting debate: similarly to neighbor-based, clustering algorithms rely on distance functions to estimate if a data point has to be assigned to a given cluster. Then, the question is: *why clustering and neighbor families result in noticeably different metric scores?* To the best of our knowledge, neighbor-based algorithms as kNN were born supervised, assigning a label to a new data point depending on the labels of their neighbors, and they were adapted lately to work in an unsupervised manner. Clustering algorithms, instead, are perfectly suited to work unsupervised without requiring any adjustment. Therefore, while they share intrinsic characteristics as the distance functions, better scores of clustering with respect to neighbor-based algorithms are consistent with the unsupervised scenario we are investigating in this paper.

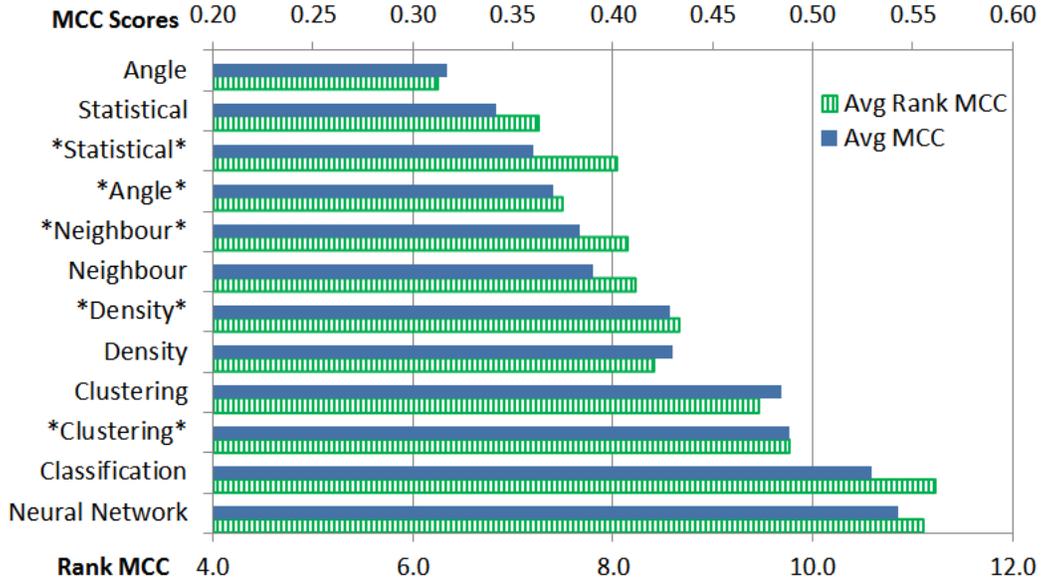

Figure 5. Average *MCC* and average *Rank MCC* for each family considered in this study, aggregated through algorithms and datasets. The * indicate groups of algorithms that belong to multiple families.

**Classification and Neural Networks.** Wrapping up the section, ***Classification (especially One-Class SVM) and SOM (Neural Network) algorithms are the most promising choices***, *while Clustering algorithms as LDCOF or G-Means may represent a good alternative when performance requirements are strict*. As a last remark, Figure 5 does not report error bars with standard deviation. We avoided error bars to increase readability of the picture, considering that standard deviation for MCC is between 0.31 and 0.33 for all the families, and that standard deviation of MCC does not provide useful hints for discussion.

## 5 Discussion on Datasets and Features

Another dimension of our analysis is related to the datasets used in our study. We selected 11 datasets, each containing data related to both normal traffic and a set of attacks, reported in separate log files: this means 51 couples <dataset, attack>. This provides a view on i) which features are selected for each dataset, and ii) how algorithms are able to detect anomalies generated by attacks in a specific dataset.

### 5.1 Selected Features

Before discussing algorithms results on each dataset, we detail the structure of each dataset and its features. The structure of the datasets constitutes a baseline for all algorithms, as it defines how data is structured and the amount of features that can be used by algorithms to detect anomalies. It is worth mentioning that particularly good or bad *detection scores of algorithms on a given dataset may be partially explained from the amount and the characteristics of datasets features*.

Features may either be numerical or textual: textual features are usually disregarded as they i) need a dedicated preprocessing depending on their intrinsic meaning, or ii) are unusable at all e.g., the "*payloadAsBase64*" feature of ISCX12, which reports the encrypted full payload of the packet. Nevertheless, numeric features may not be useful as well. In some cases, numerical features represent categories e.g., socket port, which are not meaningful for all algorithms. As example, if algorithms are based on distance functions, they are going to evaluate port 80 as "close" to port 81 and very far to e.g., port 443: calculating distance between values may be meaningless.

Table 2. Datasets details: tag, name, year and number of attacks, along with details on features as the total amount, the number of numeric features, and the number of ordinal numerical features. Selected ordinal features are reported along with their average InformationGain in the last columns on the right.

| Tag | Dataset Name | # Attacks | # Features Total | Numeric | Ordinal | Selected Features | Avg InfoGain |
|---|---|---|---|---|---|---|---|
| NF | Netflow_IDS | 3 | 11 | 5 | 5 | bytes, packets, Duration | 0.236 |
| AM | AndMal17 | 4 | 85 | 80 | 77 | act_data_pkt_fwd, Fwd Header Length 1, Fwd Header Length 2 | 0.005 |
| C7 | CICIDS17 | 5 | 85 | 80 | 77 | Packet_Length_Mean, Init_Win_bytes_backward, Packet_Length_Variance | 0.500 |
| C8 | CICIDS18 | 6 | 85 | 80 | 77 | Init_Fwd_Win_Byts, Bwd_Pktss, Fwd_Seg_Size_Min | 0.590 |
| CI | CIDDS | 4 | 16 | 7 | 5 | Bytes, Duration, Packets | 0.280 |
| CT | CTU13 | 1 | 16 | 8 | 6 | Dur, SrcBytes, dTos | 0.006 |
| IX | ISCX12 | 4 | 16 | 6 | 4 | totalDestinationPackets, totalDestinationBytes, totalSourceBytes | 0.571 |
| NG | NGDIS | 7 | 9 | 3 | 2 | event, process | 0.276 |
| NK | NSLKDD | 4 | 42 | 37 | 37 | same_srv_rate, symbolic, src_bytes | 0.607 |
| UG | UGR16 | 5 | 13 | 7 | 4 | Byte_Exchanged, Pkt_Exchanged, Serv | 0.125 |
| UN | UNSW | 8 | 45 | 39 | 38 | sbytes, sload, smean | 0.435 |

Consequently, in Table 2 we reported, for each dataset, the total amount of features, the number of numeric features and the number of ordinal numeric features. A detailed analysis of each single feature of each dataset, which may help translating some textual or categorical values into ordinal, is out of scope in this paper; therefore we consider only ordinal features of each dataset. AndMal, CICIDS17 and CICIDS18 datasets are collected with the same methodology, and report on the exact same features, which are also the larger amount (85, 77 ordinal) out of the datasets selected in our study. Indeed, there are datasets which contain just a few ordinal features i.e., Netflow-IDS, CTU13, CIDDS, ISCX12, NGDIS, UGR16, meaning that during feature selection we are not going to have many possible choices.

After selecting the best 3 ordinal features according to the feature selection strategy Information Gain, we obtain the features in the "Selected Features" column of Table 2, for each dataset. While selected features are different for all datasets, some trends can be observed. The *duration of the connection* and the *amount of bytes* exchanged, together with the *number of packets*, were selected in the majority of the datasets. Noticeably, selected features for AndMal, CICIDS17 and CICIDS18 are different, despite these datasets share the same pool of available features. This is related to the way attacks impact feature values in different datasets, which make a single feature more relevant in a dataset and less relevant in another context. Ultimately, the last column of the table reports on the average *InformationGain* score (best is 1, worst is 0) that was obtained by the selected features. This score gives an indication of the relevance of the features: briefly, *if selected features average an InformationGain score that is close to 0, we expect algorithms to raise more FPs and FNs* with respect to datasets in which these average scores are higher.

## 5.2 Detection Capabilities for each Dataset

Table 3 provides average and standard deviation for some of the common metrics based on confusion matrix. Metric values refer to algorithms scores, which are aggregated depending on the dataset they were processing. This view allows understanding *if and how algorithms encountered more difficulties in detecting attacks in a given dataset*. In particular, the last two columns i.e., Best MCC, of the table refer to the MCC that was achieved by the best algorithm when detecting a given attack on a dataset. All datasets but CTU13 report on more than one attack, and therefore these values are reported as average and standard deviation among all the attacks in this specific dataset.

Table 3. Metric Scores of all algorithms on each dataset.

| # | Dataset | TNR | P | R | F1 | | F2 | ACC | | MCC | | Best MCC | |
|---|---------|-----|---|---|----|----|----|-----|----|-----|----|----------|----|
| | | Avg | Avg | Avg | Avg | Std | Avg | Avg | Std | Avg | Std | Avg | Std |
| NF | Netflow-IDS | 0.892 | 0.72 | 0.93 | 0.74 | 0.24 | 0.80 | 0.90 | 0.06 | 0.75 | 0.26 | 0.89 | 0.20 |
| AM | AndMal17 | 0.665 | 0.23 | 0.37 | 0.18 | 0.05 | 0.24 | 0.62 | 0.06 | 0.05 | 0.63 | 0.10 | 0.04 |
| C7 | CICIDS17 | 0.647 | 0.47 | 0.72 | 0.47 | 0.29 | 0.53 | 0.68 | 0.19 | 0.37 | 0.23 | 0.70 | 0.38 |
| C8 | CICIDS18 | 0.806 | 0.75 | 0.76 | 0.67 | 0.18 | 0.71 | 0.73 | 0.19 | 0.59 | 0.32 | 0.84 | 0.23 |
| CI | CIDDS | 0.601 | 0.42 | 0.77 | 0.43 | 0.34 | 0.49 | 0.63 | 0.33 | 0.36 | 0.48 | 0.56 | 0.36 |
| CT | CTU13 | 0.752 | 0.03 | 0.33 | 0.03 | 0.00 | 0.05 | 0.75 | 0.00 | 0.04 | 0.16 | 0.25 | 0.00 |
| IX | ISCX12 | 0.778 | 0.66 | 0.78 | 0.63 | 0.36 | 0.65 | 0.80 | 0.15 | 0.56 | 0.17 | 0.86 | 0.16 |
| NG | NGDIS | 0.796 | 0.40 | 0.65 | 0.39 | 0.13 | 0.45 | 0.79 | 0.07 | 0.38 | 0.26 | 0.86 | 0.15 |
| NK | NSLKDD | 0.875 | 0.52 | 0.55 | 0.44 | 0.17 | 0.46 | 0.86 | 0.10 | 0.41 | 0.53 | 0.66 | 0.07 |
| UG | UGR16 | 0.699 | 0.44 | 0.65 | 0.37 | 0.29 | 0.39 | 0.67 | 0.23 | 0.33 | 0.29 | 0.51 | 0.28 |
| UN | UNSW | 0.853 | 0.73 | 0.56 | 0.55 | 0.16 | 0.54 | 0.80 | 0.13 | 0.47 | 0.36 | 0.70 | 0.12 |

Best MCC is on average very low for AndMal17 and – to a lesser extent – CTU13. In the general case, this is due either to i) attacks which are very similar to the normal behavior, or ii) features that are not informative enough to allow algorithms detecting attacks. We are not able to conclude anything with respect to the former item since we did not control the way attacks were conducted against the system. A rough estimation of the latter item above is provided by the last column of Table 3, which reports on average InformationGain scores of the selected features. It is interesting to see how AndMal17 and CTU13 offer the worst *InformationGain* scores of all the 11 datasets, indicating that available features do not contain sufficient information to allow algorithms to detect attacks effectively.

Instead, it is worth observing how *datasets as Netflow-IDS, ISCX12 and NGDIS contain attacks which are easier to identify*. The average value of Best MCC is the highest among all the datasets, meaning that it was possible to find a set of algorithms which detected effectively all the attacks in the dataset. Netflow-IDS reports also an high average MCC, meaning that almost all the algorithms did not raise many FPs and FNs when detecting anomalies due to attacks in that dataset. This trend does not scale for ISCX12 and NGDIS, which report average MCC of 0.56 and 0.38. In this case, the choice of the algorithm is critical since there are big differences in detection capabilities of the algorithms i.e., average MCC is lower than average value of Best MCC.

## 6 In-Depth View on Attacks

After examining algorithms and datasets, we discuss below on the detection of individual attacks and of attack categories as defined by threat landscape reports presented in Section 2.1. Furthermore, we examine how the detection capabilities of algorithms vary if they have to detect anomalies from all the attacks considered in a given dataset. This last aspect is explored in Section 6.3.

### 6.1 Categories of Attacks

We start this discussion by considering categories of attacks as defined in the ENISA report [11], namely Malware, WebAtt, WebApp, Spam/Phishing, Dos, BotNet, and DataBreach. To such extent, Figure 6 depicts a bar chart with MCC, F1, ACC and Best MCC that we obtained by aggregating algorithms scores in detecting attacks belonging to a given category, across all datasets considered in our study. We keep considering MCC as reference metric, while reporting F1 and ACC for comparison with other studies. In this case, fluctuations of MCC are coupled with similar fluctuations of F1: (D)DoS, WebAtt and, to a lesser extent, WebApp, resemble attacks that are on average identified with less errors, either FPs or FNs. We expected DoS attacks to be easy to identify by looking at network traffic, as well as web attacks such as Brute-Force. Instead, attacks as Probing/Reconnaissance, which aim at gathering sensitive data through passive scanning, usually hides into the usual traffic, and therefore may be tricky to identify.

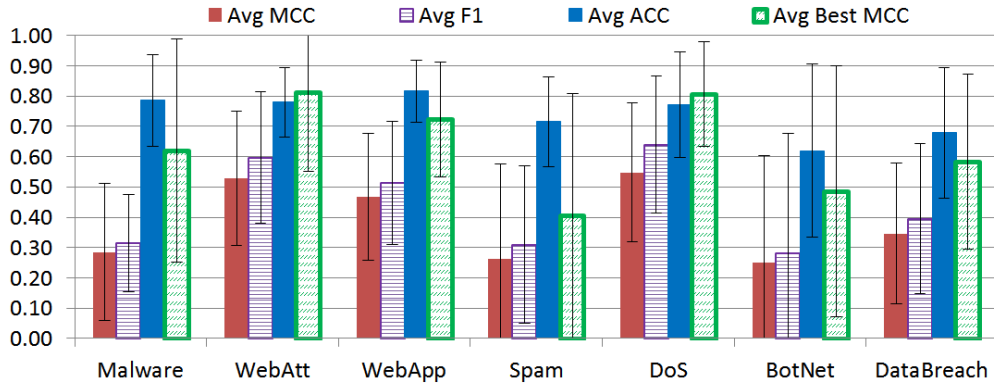

Figure 6. Average and standard deviation MCC, F1, ACC, Best MCC of algorithms in detecting attacks, grouped and ranked from left to right according to the ENISA categories reported in Section 2.1.

## 6.2 A Focus on Specific Attacks

We analyze if and how specific attacks contribute to the aggregate scores we reported in the previous section. Therefore, *Figure 7 reports a bar chart which is structured as Figure 6, focusing on specific attacks belonging to each category*. From left to right of Figure 7, we can observe two subcategories of Malware: i) Worms, and ii) Other Malware. The difference in the average capability of algorithms in detecting these two subcategories is remarkable: *worms are far more difficult to detect with respect to generic malwares*. This is enforced by the green pattern-filled bars representing the average MCC obtained by the optimal algorithm on each dataset that reports either on Worms or Other Malware. While for generic malware in each dataset it is possible to find algorithms that reach on average an MCC of 0.88, when detecting anomalies due to worms the average MCC is 0.42. A possible takeaway is that the detection of worms needs additional data about the system e.g., disk accesses, system calls, which is not usually provided in datasets which report on network traffic. The behavior of worms and the damage they can generate is not always related to network activity: as a result, our algorithms are not able to detect them since the datasets exclusively report on network features.

Differences can be observed also by looking at two different Web Attacks: Bruteforce and Fuzzers. While the former is easier to identify, the latter creates more difficulties to algorithms. Fuzzing relies on providing invalid, unexpected, or random data as network packets to a system: *its intrinsic randomness is most likely the reason why algorithms are not able to trace a clear boundary between the expected behavior of the system and anomalies due to fuzzing*, resulting in poor detection scores. A similar process works for WebApp subcategories: Backdoor and Other WebApp. The way backdoors are simulated in NGIDS and UNSW datasets is similar and embeds network data exchange, while the remaining attacks i.e., exploits, shellcode, r2l have more variability and consequently result in lower F1, ACC and MCC scores with respect to the detection of backdoors.

Figure 7. Average MCC, F1, ACC, Best MCC of algorithms in detecting specific attacks. The category of each specific attack is reported as secondary label in the footer of the column chart.

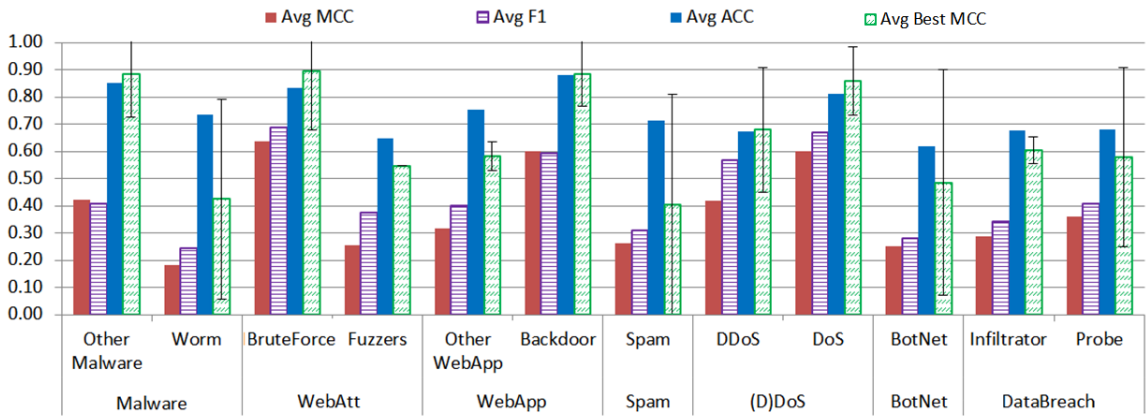

The last noticeable differences that Figure 7 highlights are related to DoS attacks. In particular, *distributed DoS attacks are trickier to identify with respect to its non-distributed variant*. The usage of different machines or different vectors to conduct the attack is employed exactly to confuse the entity – either software or a human – which is responsible of detecting these attacks. Similar detection capabilities are achieved by algorithms in detecting DataBreaches attacks, while we could not identify subcategories of Spam (datasets CTU13, UGR16, CICIDS18) and BotNet (datasets CTU13, UGR16, Netflow-Ids) categories since the way they are respectively simulated in datasets is indeed very similar.

## 6.3 Detecting all Attacks in a Dataset

In the previous sections we trained each unsupervised algorithm on each portion of each dataset separately, calculating metric scores of each algorithm in identifying a given attack without relying on labels. Instead, **we discuss here if and how training algorithms with a mixture of normal data points and data points related to different attacks of each dataset** impacts their detection capabilities. *While datasets are usually shared as an ensemble of different files that report on normal data and data related to specific attacks, when putting an IDS in a real context we do want our algorithms to detect all the possible threats*. As a result, we trained the 17 algorithms once for each of the 11 datasets, and the same model obtained after training is used to perform intrusion detection in a validation set of approximately 10.000 data points which includes data related to all the attacks reported in a given dataset.

Results of this last analysis are reported in Table 4, for each dataset. For easiness of comparison, the table is partitioned into 3 blocks. The block on the left reports average F1, ACC, MCC and Best MCC values as in Table 3. The second block i.e., unknowns, reports on average F1, ACC and MCC achieved by algorithms on the validation set which includes all attacks in a given dataset, while the last block on the right provides insights of the optimal algorithm for each dataset, i.e., the algorithm that shows the highest MCC in detecting all types of attacks in a dataset.

Table 4. F1, ACC and MCC metric scores for i) the average of all algorithms on all the attacks in the dataset, ii) the average of all algorithms, trained once for dataset, and iii) the metric scores obtained by the best algorithm in detecting unknowns, alongside with parameters values for reproducibility.

| Dataset | Avg <Dataset, Attack> | | | Avg Unknowns | | | Best Algorithm - Unknowns | | | | |
|---|---|---|---|---|---|---|---|---|---|---|---|
| | F1 | ACC | MCC | F1 | ACC | MCC | F1 | ACC | MCC | Best Algorithm (MCC) | Parameter Values |
| Netflow_IDS | 0.74 | 0.90 | 0.75 | 0.84 | 0.96 | 0.83 | 0.93 | 0.98 | 0.91 | SVM | Kernel: RBF, nu:0.02 |
| AndMal17 | 0.18 | 0.62 | 0.05 | 0.17 | 0.68 | 0.03 | 0.25 | 0.84 | 0.06 | IS_FOREST | Trees: 5, samples:20 |
| CICIDS17 | 0.47 | 0.68 | 0.37 | 0.87 | 0.92 | 0.82 | 0.90 | 0.95 | 0.87 | FASTABOD, SOM | K:5 min_a:0.1, base_a:0.6, decay:0.9 |
| CICIDS18 | 0.67 | 0.73 | 0.59 | 0.78 | 0.84 | 0.72 | 0.90 | 0.99 | 0.89 | SOS | H:50 |
| CIDDS | 0.43 | 0.63 | 0.36 | 0.70 | 0.90 | 0.69 | 0.86 | 0.96 | 0.80 | DBSCAN, SVM | min_pts:2, eps:100 Kernel: LINEAR, nu:0.02 |
| CTU13 | 0.03 | 0.75 | 0.04 | 0.03 | 0.89 | 0.04 | 0.22 | 0.99 | 0.25 | IS_FOREST | Trees: 5, samples:50 |
| ISCX12 | 0.63 | 0.80 | 0.56 | 0.82 | 0.83 | 0.71 | 0.82 | 0.97 | 0.84 | KMEANS, SOM | K:2 min_a:0.1, base_a:0.6, decay:0.9 |
| NGDIS | 0.39 | 0.79 | 0.38 | 0.35 | 0.84 | 0.37 | 0.87 | 0.99 | 0.86 | SVM | Kernel: RBF, nu:0.01 |
| NSLKDD | 0.44 | 0.86 | 0.41 | 0.61 | 0.71 | 0.40 | 0.74 | 0.79 | 0.57 | SOM | min_a:0.1, base_a:0.6, decay:0.9 |
| UGR16 | 0.37 | 0.67 | 0.33 | 0.01 | 0.94 | 0.02 | 0.12 | 0.97 | 0.18 | ISOS | K:20, phi:0.1 |
| UNSW | 0.55 | 0.80 | 0.47 | 0.35 | 0.93 | 0.40 | 0.62 | 0.95 | 0.60 | GMEANS | - |

We first focus on the last column of Table 4, which shows the algorithm(s) that achieved the highest MCC for each dataset. **SVM (3), Isolation Forests (2) and SOM (3) are the algorithms which appear the most**, while 3 clustering **algorithms as DBSCAN, K-Means and G-Means make their appearance** in three different datasets. We already debated on how Neural Network, Classification and Clustering families usually have the lowest error – either FP or FN – rates amongst all the families considered in this study. However, the most important takeover of Table 4 is that **the detection capabilities of algorithms do not degrade with large pools of attacks**. Scores of the first block of Table 4 are comparable with the scores in the other two blocks, and in some cases the results are even better. *Here relies the strength of unsupervised algorithms*. Their training never relies on labels; therefore the expected behavior of a system is reconstructed without a precise definition of attacks that may impact the system. The amount of FPs and FNs they raise is higher than supervised algorithms: therefore, *supervised algorithms should be preferred when an attack is previously known*. However, **unsupervised algorithms cover against zero-day attacks with a good scores** i.e., MCC values under Best – Algorithm in Table 4 are higher than 0.8 in 6 out of the 11 datasets, **and therefore represent a critical asset to protect a system from unknowns**.

## 7 Concluding Remarks

To conclude the paper, we report in this section threats to the validity of our study, we provide a discussion on the arising trend of attacking anomaly-based Intrusion Detectors, which is being noted and reported in the last months [48] and may become relevant in the future. Then, we summarize the findings of the paper, which finalize and wrap up our study.

### 7.1 Limitations to Validity

We report here possible limitations to the validity and the applicability of our study. These are not to be intended as showstoppers when considering the conclusions of this paper. Instead, they should be interpreted as boundaries or possible future implications which may impact the validity of this study.

**Usage of Public Data.** The usage of public data and public tools to run algorithms was a prerequisite of our analysis, to allow reproducibility and to rely on proven-in-use data. However, the heterogeneity of data sources, their potential lack of documentation and the means the authors used to collect data may limit the understandability of data. In addition, such datasets are not under our control: therefore, possible actions as changing the way data is generated by considering more features to improve detection scores are out of consideration. As for example, this could have been useful for NGDIS dataset, which has just 2 ordinal numeric features.

**Algorithms Parameters.** Each algorithm relies on its own parameters. Finding the optimal values of parameters is a substantial process which requires sensitive analyses and is directly linked with the target system in which the algorithm is going to be exercised. When applying different algorithms to different datasets it is not always possible to perform dedicated tuning of these parameters, while in other cases this activity requires perfect understanding of the insights of an algorithm and/or the target system. Results discussed in this study are obtained by using default parameter setup of the supporting tool, which embeds some parameters combinations for each algorithm, and automatically choose the most suitable for a given dataset. As a consequence, algorithms may show metric scores that still have room for improvement when using adequate parameters, if the optimal combination is not being considered by the tool.

**Variability of the Threat Landscape.** The current threat landscape is continuously mutating to include novel threats or to match updates of existing threats, which may evolve to spread rapidly or to change the way they impact a system. While attacks commented in this study are related to studies in the last decade, and are classified according to the most recent threat classification, it is straightforward that the frequency and some details of attacks are going to change in the next years. Novel attacks may be discovered, and/or the frequency of common attacks as of now may decrease, limiting their role in the overall landscape. However, observing how systems and attacks evolved in the last years, it is possible to observe continuity in the relevance of attacks e.g., Malware or DoS, which represent key threats to systems since decades. Hence, we do not expect the relevance of our study to rapidly decrease throughout years.

## 7.2 An Arising Trend: Attacks to ML-based Intrusion Detectors

The widespread use of ML algorithms to detect intrusions has a growing interest also from the viewpoint of an attacker. According to [48], "[…] *2019 was the year where attacks against machine learning security systems came into their own.*" Machine learning systems have their own weaknesses, which - with some technical expertise - can be evaded in ways that are analogous to how attackers evade "conventional" malware detectors. For example, Skylight Cyber published an attack against Blackberry/Cylance's PROTECT engine[7], showing how appending a list of strings to the end of any malware could trick PROTECT's false positive suppression component into whitelisting the malware.

While this trend is not consolidated yet, in the last months there were evidences of attacks against machine learning malware detection models that are beginning to move from the theoretical space into toolkits of attackers, impacting the role of classifiers in detecting intrusions. Therefore, we expect the results of this study to be affected by future analyses on the robustness of algorithms with respect to possible weaknesses to be exploited by attackers. Alongside with misclassifications, we expect future comparisons to consider aspects related to the robustness to these reverse engineering attacks to ML-based intrusion detectors.

## 7.3 Lessons Learned

As a last contribution, we summarize the main findings of this paper as follows.
- As discussed in Section 4, **Classification algorithms** as One-Class SVM [28] and Isolation Forest [42], **as well as Self Organizing Maps** (SOM [30]), turned out to be the algorithms with less misclassifications, either FPs or FNs.
- As a side effect of the analyses above, **clustering algorithms can represent a valid alternative to classification and neural networks when performance constraints** forces the adoption of algorithms which execute training in semi-linear time, instead of quadratic as it is required by SVM, Isolation Forests and SOM.
- As expected, different types of attacks generate different anomalies, which are not detected by algorithms with the same metric scores. In particular, *Spam*, malware such as *Worms*, and attacks which rely on a coordinated and distributed action as *Distributed DoS* and *BotNets* are more likely to be misclassified by intrusion detectors.
- Amongst all the different types of attacks, **different datasets exhibit different detection scores when applying unsupervised anomaly detectors**. This is mainly due i) to their structure, namely the amount and the characteristics of features which build the dataset, and, to a lesser extent, ii) to the way authors simulated them when gathering system data.
- Applying **unsupervised algorithms to detect an heterogeneous pool of attacks does not dramatically reduce their detection capabilities** with respect to when they are requested to identify anomalies due to a specific type of attack. Section 6.3 demonstrates how, in some cases, detection capabilities are even higher in the former – and more realistic – case.

In a nutshell, when building an intrusion detector for an existing or a brand new system, our study suggests betting either on Classification i.e., One-Class SVM, Isolation Forests, or Self-Organizing Maps algorithms, to detect zero-day attacks. However, depending on the domain, betting is not an option. That is why dedicated tuning strategies and sensitive analyses of algorithms' parameters tailored to the specific system

---

[7] SkyLight Cylance, I Kill You!, https://skylightcyber.com/2019/07/18/cylance-i-kill-you/ (online)

are – and will always be – mandatory to devise intrusion detectors that effectively deal with both known and zero-day attacks in a specific system.

## ACKNOWLEDGMENTS

This work has been partially supported by the REGIONE TOSCANA POR FESR 2014-2020 SISTER and by the H2020 programme under the Marie Sklodowska-Curie grant agreement 823788 (ADVANCE) projects.